%% file: RIFAR_ACL.tex
\documentclass[11pt]{article}
\usepackage[preprint]{acl}
\usepackage{makecell}
\usepackage{algorithm}
\usepackage{algpseudocode}
\usepackage{times}
\usepackage{latexsym}
\usepackage[T1]{fontenc}
\usepackage[utf8]{inputenc}
\usepackage{microtype}
\usepackage{inconsolata}
\pdfmapfile{+zi4.map}
\usepackage{amsmath,amssymb,bm,graphicx,booktabs,array,longtable,multirow}
\usepackage{tabularx}
\usepackage{xcolor,colortbl}
\input{abstract-background}
\usepackage{enumitem}
\usepackage{placeins}
\usepackage{stfloats}
\usepackage{needspace}
\usepackage{xurl}
\input{table-style}
\setlist[itemize]{leftmargin=*} % Keep horizontal indentation; inherit vertical spacing.

\title{RIFAR: Reliability-Informed and Forgetting-Aware Replay for Continual Robot Learning with World-Action Models}
\hypersetup{
  pdftitle={RIFAR: Reliability-Informed and Forgetting-Aware Replay for Continual Robot Learning with World-Action Models},
  pdfauthor={Tu Lin, Zheng Lu, Zirui Song, Zixian Lin, Zebang Feng, Meijia Chen, Xiaoyu Sun, Wanru Li, Lijie Wang, Xiuying Chen},
  pdfsubject={Continual robot learning with world-action models},
  pdfkeywords={Continual Robot Learning, World-Action Models, Generative Replay}
}
\newsavebox{\RIFARAuthorNotes}
\author{
Tu Lin\textsuperscript{1,*},\quad
Zheng Lu\textsuperscript{2,*,\textdaggerdbl},\quad
Zirui Song\textsuperscript{3,*},\quad
Zixian Lin\textsuperscript{1},\quad
Zebang Feng\textsuperscript{1}\\[3pt]
\bfseries Meijia Chen\textsuperscript{4},\quad
Xiaoyu Sun\textsuperscript{1},\quad
Wanru Li\textsuperscript{1},\quad
Lijie Wang\textsuperscript{5,\textdagger},\quad
Xiuying Chen\textsuperscript{3,\textdagger}\\[5pt]
\normalfont\normalsize
\textsuperscript{1}South China University of Technology\quad
\textsuperscript{2}Peking University\\[2pt]
\normalfont\normalsize
\textsuperscript{3}MBZUAI\quad
\textsuperscript{4}Rutgers University\quad
\textsuperscript{5}Zhejiang University\\[4pt]
\normalfont\small
\sbox{\RIFARAuthorNotes}{%
\textsuperscript{*}Equal contribution.\quad
\textsuperscript{\textdagger}Corresponding authors.\quad
\textsuperscript{\textdaggerdbl}Project Leader.}%
\makebox[0pt][l]{\makebox[\wd\RIFARAuthorNotes][c]{%
\raisebox{-12pt}[0pt][0pt]{\normalfont\small\texttt{202621062928@mail.scut.edu.cn}}}}%
\usebox{\RIFARAuthorNotes}
}
\begin{document}
\maketitle
\begin{abstract}
\input{sections/abstract}
\end{abstract}
\input{sections/introduction_new}
\input{sections/related_work}
\input{sections/method}
\input{sections/ltt_exp}
\input{sections/conclusion}
\input{sections/limitations}

\bibliography{references}
\clearpage
\appendix
\input{sections/appendix_lt_A}
\input{sections/appendix_lt_B}

\input{sections/appendix_lt_C}

\end{document}

%% file: abstract-background.tex
\usepackage{tikz}
\definecolor{abstractpanel}{RGB}{246,249,255}
\AddToHookNext{shipout/background}{%
  \begin{tikzpicture}[remember picture,overlay]
    \fill[abstractpanel,rounded corners=2mm]
      ([xshift=80.432bp,yshift=218.833bp]current page.south west)
      rectangle
      ([xshift=280.840bp,yshift=613.428bp]current page.south west);
  \end{tikzpicture}%
}

%% file: table-style.tex
\usepackage{xcolor,colortbl,tabularx}
\definecolor{TableHeader}{HTML}{E8EEF4}
\definecolor{RifarRow}{HTML}{F4E4D3}
\definecolor{RifarInk}{HTML}{85522F}
\definecolor{TableStripe}{HTML}{F7F9FB}
\definecolor{TableRule}{HTML}{708297}
\definecolor{TableMuted}{HTML}{667788}
\definecolor{SuiteGoalFill}{HTML}{E3EEE8}
\definecolor{SuiteGoalInk}{HTML}{2F695B}
\definecolor{SuiteObjectFill}{HTML}{DFEAF5}
\definecolor{SuiteObjectInk}{HTML}{345F86}
\definecolor{SuiteSpatialFill}{HTML}{EBE5F2}
\definecolor{SuiteSpatialInk}{HTML}{66507F}
\colorlet{IDMFill}{SuiteObjectFill}
\colorlet{DriftFill}{SuiteGoalFill}
\definecolor{FixedReplayFill}{HTML}{EAF0F4}
\newcommand{\RIFARTable}{%
  \small
  \renewcommand{\arraystretch}{1.04}%
  \setlength{\tabcolsep}{4pt}%
  \setlength{\heavyrulewidth}{0.65pt}%
  \setlength{\lightrulewidth}{0.3pt}%
  \setlength{\cmidrulewidth}{0.3pt}%
  \arrayrulecolor{TableRule}%
}
\newcolumntype{C}{>{\centering\arraybackslash}X}
\newcolumntype{L}{>{\raggedright\arraybackslash}X}

%% file: sections/abstract.tex
% Genuine embodied agency requires robots to turn continuous real-world experience into lasting, transferable skills. This demands continual learning that integrates new capabilities while preserving prior knowledge in evolving environments. Experience replay mitigates catastrophic forgetting, but storing complete demonstrations becomes costly as tasks accumulate. World–action models provide an alternative through generative replay. However, visually coherent rollouts may contain actions that cannot realize the predicted transitions, while new-task adaptation can disrupt learned behavior. To address these problems, we propose \textbf{RIFAR (Reliability-Informed and Forgetting-Aware Replay)}, a reliability-informed and drift-aware generative replay framework. It reconstructs trajectories from compact historical demonstration prefixes and uses a trainable inverse-dynamics model to assess action--visual consistency. It first adapts to new tasks using current-task demonstrations and high-reliability replay, then evaluates how this adaptation changes action predictions on historical inputs and prioritizes reliable trajectories with greater action drift for further training. Across three LIBERO suites and real-world experiments, \mbox{RIFAR} outperforms the previous state of the art in WAM-based generative replay. On LIBERO-Goal, it achieves an AUC of 90.97 while retaining only 320 historical time steps per task, about 4.9\% of those retained by 50-demonstration replay.

Genuine embodied agency requires robots to turn continuous real-world experience into lasting, transferable skills. This demands continual learning that integrates new capabilities while preserving prior knowledge in evolving environments. Experience replay mitigates catastrophic forgetting, but storing complete demonstrations becomes costly as tasks accumulate. World–action models provide an alternative through generative replay. However, visually coherent rollouts may contain actions that cannot realize the predicted transitions, while new-task adaptation can disrupt learned behavior. To address these problems, we propose \textbf{RIFAR (Reliability-Informed and Forgetting-Aware Replay)}, a reliability-informed and drift-aware generative replay framework. It reconstructs trajectories from compact historical demonstration prefixes and uses a trainable inverse-dynamics model to assess action--visual consistency. It first adapts to new tasks using current-task demonstrations and high-reliability replay, then evaluates how this adaptation changes action predictions on historical inputs and prioritizes reliable trajectories with greater action drift for further training. Across three LIBERO suites and real-world experiments, \mbox{RIFAR} outperforms the previous state of the art in WAM-based generative replay. On LIBERO-Goal, it achieves an AUC of 90.97 while retaining only 320 historical time steps per task, about 4.9\% of those retained by 50-demonstration replay.

%% file: sections/introduction_new.tex
\section{Introduction}

\begin{figure*}[t]
\centering
\includegraphics[width=\linewidth,trim=0 3 0 3,clip]{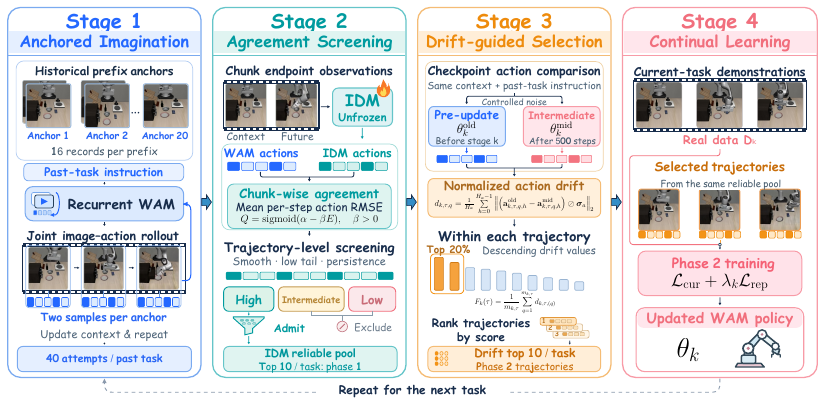}
\caption{RIFAR overview. A trainable IDM screens trajectories generated from historical prefixes. Phase 1 combines current demonstrations with each past task's ten highest-quality trajectories for 500 steps. Phase 2 reselects ten reliable trajectories per task using the mean of the largest 20\% of action drifts within each trajectory.}
\label{fig:pipeline}
\end{figure*}

Robots operating in evolving real-world environments must continually acquire new manipulation skills while retaining earlier capabilities. Yet adapting a shared action policy to new tasks can disrupt previously learned behavior, causing catastrophic forgetting \citep{kirkpatrick2017}. Effective continual learning therefore requires integrating new experience without sacrificing knowledge acquired from previous tasks over long-term deployment.

Experience replay mitigates this problem by training on stored demonstrations alongside current-task data \citep{phaser2026,infovla2026}. However, these demonstrations contain long sequences of observations, proprioceptive states, and actions, making storage increasingly costly as tasks accumulate. Generative replay reduces this storage burden by reconstructing historical training examples from model knowledge \citep{shin2017}. World--action models (WAMs) are suited to this approach: by jointly predicting actions and future observations \citep{zhu2025}, they can recurrently reconstruct past-task trajectories for replay. Despite this potential, continual learning with WAMs remains largely unexplored. REGEN takes an initial step by generating replay trajectories from historical task instructions and current-task observations for continual learning \citep{regen2026}. However, generating trajectories that effectively preserve prior knowledge remains challenging.

Two key limitations arise in WAM-based generative replay. First, REGEN's main configuration pairs a past-task instruction with a current-task observation, although the current scene may lack the objects or spatial configuration required for the earlier task. Compact historical context helps provide a more suitable starting point, but errors can still arise during recurrent generation: even a visually coherent rollout may contain actions that fail to produce the predicted visual transitions. Second, even when generated trajectories are reliable, reliability alone does not indicate which replay is most needed. As the policy learns a new task, its action predictions on identical historical inputs may change substantially, potentially disrupting previously successful manipulation behavior on some historical trajectories. Therefore, effective replay selection requires identifying trustworthy generated trajectories for replay and further prioritizing those reliable historical trajectories that are more strongly affected by new-task adaptation.

To advance embodied agency through continual learning, we develop \textbf{RIFAR} (\textbf{Reliability-Informed and Forgetting-Aware Replay}) for WAMs, supporting the acquisition of new skills and the retention of earlier capabilities (Figure~\ref{fig:pipeline}). It initializes recurrent generation of past-task trajectories from just 320 retained historical time steps per task, approximately 4.9\% of those retained by 50-demonstration experience replay. A trainable inverse-dynamics model assesses whether the generated actions are consistent with the predicted visual transitions. The policy first trains on current-task demonstrations and the highest-quality trajectories that pass screening. To guide the second phase, RIFAR compares action predictions on identical historical inputs before and after this initial adaptation, then prioritizes reliable trajectories with larger action drift. This focuses replay on earlier behaviors most affected by adaptation while training on the current task continues. We assess retention using success rates on previously learned tasks.

We evaluate RIFAR using Cosmos Policy as the backbone on LIBERO-Goal, LIBERO-Object, and LIBERO-Spatial. RIFAR consistently outperforms REGEN, improving AUC by 23.17, 5.39, and 9.93 points on the three suites, respectively. The comparison with real-data replay also highlights its favorable trade-off between retained historical data and performance. On LIBERO-Goal, RIFAR achieves an AUC of 90.97 while retaining only 320 historical time steps per task. In comparison, ten-demonstration experience replay ER(10) reaches 90.70 AUC with approximately 1,300 retained steps per task. RIFAR thus achieves slightly higher AUC while retaining approximately 75\% fewer historical time steps. This advantage over generative replay also extends to real-world manipulation: across three sequentially introduced tasks, RIFAR achieves 57.78 AUC, exceeding REGEN by 7.39 points. Ablation studies and action-drift analyses further support the complementary contributions of reliability screening and adaptive replay selection to preserving earlier manipulation skills.

Our main contributions are as follows:

\begin{itemize}[itemsep=1pt,parsep=0pt,topsep=2pt,partopsep=0pt]
\item \textbf{Reliable generative replay for WAMs.} RIFAR reconstructs past-task trajectories from compact context and uses a trainable inverse-dynamics model to reject trajectories whose actions are inconsistent with depicted visual transitions.
\item \textbf{Forgetting-aware replay selection.} RIFAR measures action drift on identical historical inputs before and after initial adaptation to identify behaviors more strongly affected by new-task learning. It then prioritizes reliable trajectories with larger drift for second-phase replay.
% \item \textbf{Valid Experiments.} Experiments on LIBERO-Goal and LIBERO-Object demonstrate substantial improvements over REGEN, while ablations isolate the contribution of each component and storage analysis shows strong performance with only 320 historical time steps retained per task.
\item \textbf{Empirical validation and storage efficiency.}
RIFAR consistently outperforms REGEN across three LIBERO suites and real-world tasks, while ablations verify each component and storage analysis shows strong performance with only 320 historical time steps retained per task.
\end{itemize}

%% file: sections/related_work.tex
\section{Related Work}

\subsection{Continual Learning}

Continual learning aims to acquire new capabilities while retaining prior knowledge. EWC constrains updates according to Fisher-based parameter importance \citep{kirkpatrick2017}, while PackNet preserves task-specific weights through iterative pruning and masking \citep{packnet2017}. LoRA reduces trainable parameters through low-rank updates to frozen pretrained weights \citep{lora2021}, but parameter efficiency alone does not guarantee retention. Experience replay revisits earlier tasks by mixing past examples with current-task data \citep{libero2023}. Complete manipulation demonstrations contain sequences of images and actions. Generative replay instead reconstructs historical training examples \citep{shin2017}, reducing reliance on stored demonstrations.

\subsection{Vision-Language-Action Models for Robot Manipulation}

Recent work applies continual learning to vision-language-action (VLA) policies \citep{lu2026layerroute}. CLARE expands adapters using feature novelty and routes inputs without task labels or stored exemplars \citep{clare2026}. For continual reinforcement learning, CRL-VLA combines goal-conditioned value modeling with asymmetric regularization \citep{crlvla2026}, while Simple Recipe Works finds limited forgetting when adapting pretrained VLAs with LoRA and on-policy reinforcement learning \citep{recipe2026}. Replay mitigates forgetting in real-world VLA experiments \citep{zhu2026can}. PHASER allocates replay capacity by manipulation phase and prioritizes samples through language--vision similarity and action divergence \citep{phaser2026}; Info-VLA uses replay-anchored contrastive learning with cross-modal mutual-information constraints \citep{infovla2026}.

% World--action models jointly predict actions and future observations, enabling them to recurrently generate complete manipulation trajectories \citep{wang2026acting,kim2026}. This capability provides a natural basis for generative replay, as earlier experience can be reconstructed by the model rather than stored in full. Despite this potential, continual learning with WAMs remains largely unexplored. REGEN represents an initial approach, training on generated trajectories from earlier tasks alongside current-task data \citep{regen2026}. To reduce historical storage, its main configuration conditions earlier-task instructions on current-task observations. This may start generation from a scene that lacks the objects or spatial configuration required by the earlier task. Moreover, a coherent predicted video does not guarantee that its paired actions would produce the depicted transitions during robot execution. Although REGEN documents these inconsistencies, it does not explicitly remove unreliable trajectories before replay or prioritize trajectories according to how strongly new-task learning changes earlier action predictions.

\subsection{World Action Models for Continual Robot Learning}

World--action models jointly predict actions and future observations, enabling them to recurrently generate complete manipulation trajectories \citep{wang2026acting,kim2026}. This capability provides a natural basis for generative replay, as earlier experience can be reconstructed by the model rather than stored in full. Despite this potential, continual learning with WAMs remains largely unexplored. REGEN represents an initial approach, training on generated trajectories from earlier tasks alongside current-task data \citep{regen2026}. However, REGEN does not explicitly screen generated trajectories for action--visual consistency or adapt replay priorities to policy changes induced by new-task learning. Replay may thus reinforce generation errors and overlook earlier behaviors at greater risk of forgetting. These limitations suggest that effective replay depends on both the quality of reconstructed experience and its value for preserving earlier skills. Ensuring reliable supervision while directing replay toward behaviors most affected by learning is therefore important for continual adaptation with WAMs, motivating RIFAR.

%% file: sections/method.tex
\section{Method}
\label{sec:method}

RIFAR combines reliability screening with adaptive replay selection for continual robot learning with WAMs. As shown in Figure~\ref{fig:pipeline}, it reconstructs past-task trajectories from compact historical contexts and screens them for action--visual consistency to form a reliable replay pool. Training proceeds in two phases. We first adapt the policy using current-task data and the highest-quality screened trajectories. We then compare its action predictions before and after this adaptation on identical historical contexts from the same replay pool. Trajectories with the largest normalized action drift are prioritized for replay in the second phase, while training on the current task continues.

\begin{figure*}[!t]
\centering
\includegraphics[width=\linewidth]{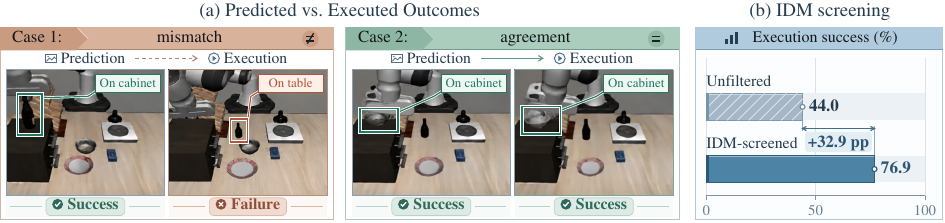}
\caption{IDM reliability screening.
(a) Predicted outcomes and simulator executions: one case fails despite predicted success, whereas the other is consistent.
(b) Execution success is 44.0\% among all unfiltered generated trajectories and 76.9\% among all IDM-screened trajectories, a difference of 32.9 percentage points.}
\label{fig:idm_case}
\end{figure*}

\subsection{Problem Setup}

We study continual robot learning, where a single policy learns $N$ manipulation tasks, denoted by $T_0,\ldots,T_{N-1}$, over successive stages. For each task $T_i$, a language instruction $\ell_i$ describes the goal, and a demonstration dataset $\mathcal{D}_i$ provides example trajectories for completing the task. Each demonstration trajectory records a sequence of visual observations, proprioceptive states, and robot actions.

After initialization, each incremental stage introduces a single new manipulation task. We denote the introduction stage of task $T_i$ by $g(i)$ and the demonstration dataset of the new task at stage $k$ by $\mathcal{D}_k^{\mathrm{new}}$. For historical tasks with $g(i)<k$, complete demonstrations are no longer available; we retain only compact historical contexts from earlier demonstrations for replay generation. Before adaptation, we save the previous policy checkpoint as $\theta_k^{\mathrm{old}}=\theta_{k-1}$ to measure subsequent action drift on identical past-task contexts. The policy is then trained on new-task demonstrations together with generated historical replay that passes screening. After each stage, we evaluate the updated policy on the new task and all previously learned tasks to assess both task acquisition and skill retention.

\subsection{Initial-State Anchored Historical Imagination}

Previous WAM-based generative replay reconstructs past-task trajectories by jointly predicting actions and future observations. Without historical context, however, generation may begin from a scene that lacks the objects or spatial configuration required by the task. We therefore introduce \textbf{Initial-State Anchored Historical Imagination}, which retains short sequences of observations and proprioceptive states from the beginning of selected historical demonstrations. These sequences provide task-compatible starting points for generation without requiring complete demonstrations to be stored. For each historical task $T_i$, we select the first $K$ demonstrations in dataset order and retain only the initial $H_s$ time steps of each selected demonstration. These short sequences form the anchor set $S_i=\{s_{i,j}\}_{j=1}^{K}$. Each anchor is defined as $s_{i,j}=\{(o_t,p_t)\}_{t=0}^{H_s-1}$, where $o_t$ and $p_t$ denote the visual observation and proprioceptive state at time step $t$ in the corresponding demonstration. We use $K=20$ and $H_s=16$, storing only $KH_s=320$ observation--proprioception pairs per task. During generation, the WAM takes the current observation $o_t$, proprioceptive state $p_t$, and task instruction $\ell_i$ as input and jointly predicts an $H$-step action chunk and the corresponding future state at $t+H$:
\begin{equation}
\textstyle
(\widehat{A}_t,\widehat{o}_{t+H},\widehat{p}_{t+H})
=
G_{\theta}(o_t,p_t,\ell_i;\xi),
\end{equation}
where $\widehat{A}_t=(\widehat{a}_t,\ldots,\widehat{a}_{t+H-1})
\in\mathbb{R}^{H\times d}$ contains $H$ consecutive action predictions,
each of dimension $d$. The outputs $\widehat{o}_{t+H}$ and
$\widehat{p}_{t+H}$ are the predicted visual observation and proprioceptive
state at time $t+H$. Here, $\theta$ denotes the WAM
parameters, and $\xi$ represents sampling randomness.

Generation begins with a retained anchor, whose observation--proprioception pairs are supplied to the WAM in temporal order. At each generation step, the WAM predicts an action chunk with a future pair. Only the first action $\widehat{a}_t$ is retained for the current step, while the predicted pair is stored for later use. The WAM then advances to the next retained pair and predicts again. Once the stored sequence is exhausted, 
the stored predicted pairs are recursively used as inputs, extending the short context into a long-horizon replay trajectory.

We denote the $b$-th generated rollout from the retained anchor $s_{i,j}$ by $\widehat{\tau}_{i,j,b}=\{(\widehat{o}_t,\widehat{p}_t,\widehat{a}_t)\}_{t=0}^{L}$, where $L$ is the final time index. All rollouts generated from the anchors of historical task $T_i$ form the candidate replay pool $\mathcal{P}^{\mathrm{cand}}_i=\{\widehat{\tau}_{i,j,b}\}_{j,b}$ for subsequent IDM-based reliability screening.

\subsection{Reliability Screening}

Although historical anchors provide useful contexts, generated actions still deviate from predicted visual transitions, as illustrated in Figure~\ref{fig:idm_case}(a). To assess this consistency, we use an inverse-dynamics model (IDM) $I_{\phi}$ that combines a trainable DINOv2 ViT-S/14 visual encoder with a two-layer Transformer decoder. Given the paired initial and endpoint images, the IDM predicts a 16-step chunk of seven-dimensional actions. Initially, we train it on LIBERO Tasks 0-5 demonstrations, using image pairs 16 time steps apart as visual inputs and the intervening action sequence as supervision. After base training, the IDM is continually updated as new tasks are introduced during continual learning.

For each generated chunk $r$, let $\widehat{o}_r$ and $\widehat{o}^{+}_r$ denote its initial observation and predicted endpoint observation, respectively. The IDM infers the action chunk as $\widehat{A}^{\mathrm{inv}}_r=I_{\phi}(\widehat{o}_r,\widehat{o}^{+}_r)$. We normalize both the IDM-inferred and WAM-generated actions using the same transformation $\widetilde{a}=\mathcal{N}(a)$ and compute their chunk-level disagreement as:
\begin{equation}
\textstyle
E_r^{\mathrm{IDM}}
=
H_a^{-1}d^{-1/2}
\sum\nolimits_{h=0}^{H_a-1}
\bigl\|
\widetilde{a}^{\mathrm{inv}}_{r,h}
-
\widetilde{a}_{r,h}
\bigr\|_2.
\label{eq:idm_error}
\end{equation}
Here, $H_a$ is the number of actions in each compared chunk, and $d$ is the action dimension. At position $h$ in chunk $r$, $\widetilde{a}^{\mathrm{inv}}_{r,h}$ and $\widetilde{a}_{r,h}$ denote the normalized IDM-inferred and WAM-generated actions, respectively. Thus, $E_r^{\mathrm{IDM}}$ measures the root-mean-square error across action dimensions at each step, averaged over the chunk. We convert this disagreement into an agreement score:
\begin{equation}
\textstyle
Q_r^{\mathrm{IDM}}
=
\sigma(\alpha-\beta E_r^{\mathrm{IDM}}),
\quad \beta>0,
\label{eq:idm_score}
\end{equation}
where $\sigma(z)=(1+\exp(-z))^{-1}$  and $\alpha$ and $\beta$ control the offset and sensitivity of the mapping. 

We smooth chunk scores with a three-chunk trailing average and define $Q(\widehat{\tau})$ as the mean of the lowest $20\%$ of smoothed scores. Screening also rejects trajectories with persistent low agreement as shown in Appendix~\ref{app:idm_scoring}. WAM-generated actions remain the targets for replay. For each historical task $T_i$, trajectories that pass screening form the replay pool $\mathcal{P}_{k,i}$. We select the ten highest-scoring trajectories from each pool for Phase 1 and denote the selected set by $\mathcal{S}^{(1)}_{k,i}$. Starting from $\theta_k^{\mathrm{old}}$, we train the policy for 500 optimization steps using current-task demonstrations and the selected replay trajectories from all historical tasks, yielding the intermediate checkpoint $\theta_k^{\mathrm{mid}}$. The screened replay pools remain available for reselection in Phase 2.

\subsection{Drift-Based Replay Prioritization}
Although action drift is not equivalent to forgetting, it can serve as a useful signal of forgetting-related behavioral change.
To guide replay reselection in Phase 2, we compare the action predictions of the pre-update policy $\theta_k^{\mathrm{old}}$ and the intermediate policy $\theta_k^{\mathrm{mid}}$, obtained after 500 optimization steps, using identical contexts from the full screened replay pools. Both use identical observations, proprioceptive states, task instructions and  inference settings. For each context $q$ in trajectory $\widehat{\tau}$, both policies predict an $H_a$-step action chunk. With stage $k$, trajectory $\widehat{\tau}$, and context $q$ fixed, we denote their actions at chunk position $h$ by $a_h^{\mathrm{old}}$ and $a_h^{\mathrm{mid}}$. We average the normalized action distances over the chunk to obtain the context-level drift:
\begin{equation}
\textstyle
 d_{k,\tau,q}=H_a^{-1}\sum\nolimits_{h=0}^{H_a-1}
 \|(a_h^{\mathrm{old}}-a_h^{\mathrm{mid}})\oslash\sigma_a\|_2,
\label{eq:drift}
\end{equation}
where $\oslash$ denotes element-wise division and the fixed positive vector $\sigma_a\in\mathbb{R}_{+}^{d}$ provides a shared normalization scale for each action dimension.

For each trajectory $\widehat{\tau}$ with $M_\tau$ evaluated contexts, let $d_{k,\tau,(q)}$ denote its $q$-th largest context-level drift. We use $m_\tau=\operatorname{Round}_{\mathrm{tail}}(0.2M_\tau)$, with $1\leq m_\tau\leq M_\tau$, to denote the nonempty number of contexts in the upper $20\%$ tail. The forgetting-aware replay priority is their mean drift:
\begin{equation}
\textstyle
F_k(\widehat{\tau})
=
m_{k,\tau}^{-1}
\sum\nolimits_{q=1}^{m_{k,\tau}}
d_{k,\tau,(q)}.
\label{eq:tail}
\end{equation}
The upper-tail mean emphasizes localized policy changes that can be diluted by averaging over the entire trajectory. For each historical task $T_i$, we select the ten highest-priority trajectories from the same full screened pool $\mathcal{P}_{k,i}$. This set replaces the Phase 1 replay set, with possible overlap. Starting from $\theta_k^{\mathrm{mid}}$, we continue training for 300 optimization steps using current-task demonstrations and the reselected replay trajectories, yielding the updated policy $\theta_k$. The two phases together comprise 800 optimization steps per incremental task.

%% file: sections/ltt_exp.tex
\input{tables/table1}

\section{Experiments}
\label{sec:experiments}

\subsection{LIBERO Evaluation Setup}
We evaluate RIFAR on the LIBERO benchmark \citep{libero2023} using Cosmos Policy \citep{kim2026}, initialized from Cosmos-Predict2-2B. The policy receives third-person and wrist RGB observations, proprioceptive states, and language instructions, and predicts a 16-step action chunk.

Across the Goal, Object, and Spatial suites, we jointly train on Tasks 0--5 and then introduce Tasks 6--9 sequentially. We train all methods for 800 optimization steps per new task and evaluate all seen tasks after each stage. For RIFAR, we select the first 20 demonstrations of each past task in dataset order. We retain only the initial 16 time steps of each demonstration for replay generation. Both training phases combine current-task demonstrations with replay. Phase 1 uses the highest-quality IDM-screened trajectories for 500 steps. We then reselect trajectories from the full reliable pool based on action drift and continue training for another 300 optimization steps in Phase 2.

We compare RIFAR with sequential fine-tuning, EWC, PackNet, REGEN, and ER($n$), which stores $n$ complete demonstrations per historical task for replay. We report AUC, FWT, and NBT for each compared method, with metric definitions and the evaluation protocol in Appendix~\ref{app:evaluation}. 
\begin{figure*}[!b]
\centering
\includegraphics[width=\linewidth]{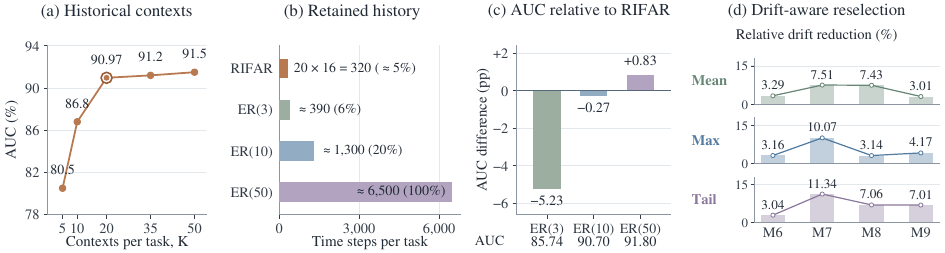}
\caption{Historical replay and action drift on LIBERO-Goal.
(a) AUC versus the number of anchors per task (default $K=20$).
(b) Retained time steps relative to ER(3/10/50).
(c) AUC differences of ER baselines relative to RIFAR (90.97).
(d) Relative drift reduction on Task 3 across M6--M9, positive values indicate reduced drift.}
\label{fig:anchors}
\label{fig:drift_reduction}
\end{figure*}

\subsection{Main Results and Storage Efficiency}

\begin{figure*}[!t]
\centering
\includegraphics[width=\linewidth]{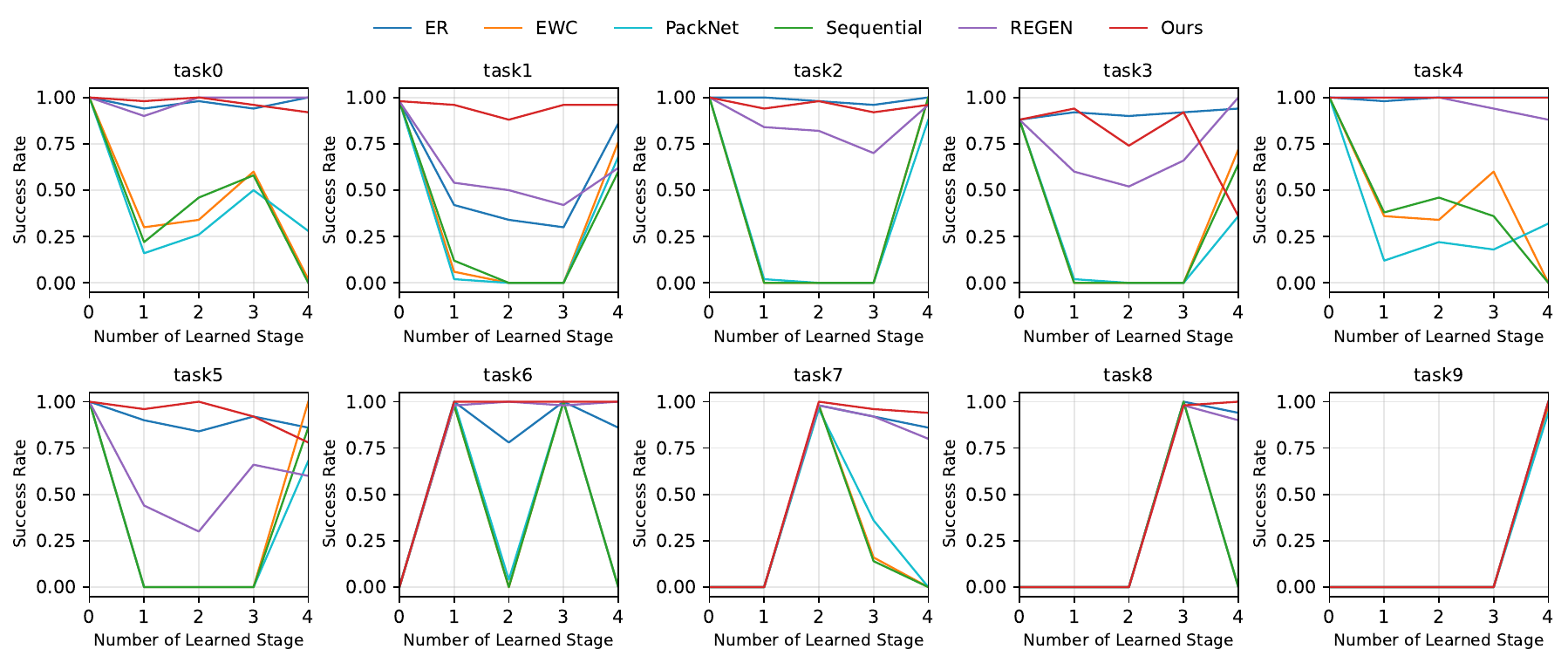}
\caption{Per-task success rate comparison on LIBERO-Object, averaged over three random seeds. ER denotes ER(3). RIFAR maintains high success rates across tasks with only minor fluctuations on a small number of tasks.}
\label{fig:object}
\end{figure*}

As shown in Table~\ref{tab:main} and Figure~\ref{fig:object}, RIFAR achieves higher AUC and lower NBT than REGEN across all three evaluated LIBERO suites. On LIBERO-Goal, RIFAR improves AUC from 67.8 to 90.97 and reduces NBT from 46.7 to 7.45 compared with REGEN. The improvements remain consistent on LIBERO-Object and Spatial, with AUC gains of 5.39 and 9.93 points over REGEN, respectively.

RIFAR also remains competitive with experience replay using stored demonstrations, while retaining only compact initial-state anchors rather than complete trajectories. As shown in Figure~\ref{fig:anchors}(b,c), we further investigate the
relationship between historical data storage and performance through a separate
study on LIBERO-Goal.  RIFAR achieves an AUC of 90.97, slightly exceeding the 90.70 obtained by ER(10) while retaining approximately one-quarter as many historical time steps. ER(50) achieves a higher AUC of 91.80 but requires approximately 6,500 time steps per task. These results show that RIFAR remains competitive with experience replay while retaining substantially less historical data.

\subsection{Ablation Studies}

% \textbf{Effect of the Number of Anchors.}
% We vary the number of anchors per historical task on LIBERO-Goal to assess the trade-off between retained context and continual learning performance (Figure~\ref{fig:anchors}(a)). AUC increases from 80.5 at $K=5$ to 86.8 at $K=10$ and 90.97 at $K=20$, then rises more slowly to 91.2 at $K=35$ and 91.5 at $K=50$. We therefore use $K=20$ as our default setting for each task, which retains 40\% as many anchors as $K=50$ with AUC only 0.53 points lower.

\textbf{Effect of the Number of Anchors.}
On LIBERO-Goal, increasing the number of anchors per historical task from 5 to 20 raises AUC from 80.5 to 90.97, as shown in Figure~\ref{fig:anchors}(a). Beyond 20 anchors, the gains become much smaller, with $K=50$ reaching 91.5 AUC. This suggests that a small set of historical contexts can support effective replay during adaptation to newly introduced tasks. We therefore use $K=20$ as the default to balance storage and performance, retaining only 40\% as many anchors as $K=50$ while keeping AUC within 0.53 points.

\input{tables/table2}

% \textbf{IDM Screening and Drift-Based Selection.}
% We compare anchor-only replay with IDM screening, drift-based selection, or both (Table~\ref{tab:ablation}). Either component improves AUC and reduces NBT. On LIBERO-Goal, adding drift-based selection to IDM screening raises AUC from 87.14 to 90.97 and reduces NBT from 10.61 to 7.45. RIFAR achieves the best AUC and NBT across all three suites.
\textbf{IDM Screening and Drift-Based Selection.}
The ablations in Table~\ref{tab:ablation} show that either component improves AUC and reduces NBT compared with anchor-only replay. On LIBERO-Goal, adding drift-based selection to IDM screening raises AUC from 87.14 to 90.97 and lowers NBT from 10.61 to 7.45. RIFAR achieves the highest AUC and lowest NBT across all three LIBERO suites.

\input{tables/table8}

\textbf{Analysis of Replay Selection Criteria.}
Table~\ref{tab:selection_validation} directly evaluates whether IDM reliability and action drift provide meaningful signals for replay selection by comparing high-ranked, randomly sampled, and low-ranked trajectories at stage 1 of each LIBERO suite. High-ranked selection yields the highest success rate in all three suites. These results support IDM reliability and action drift as effective signals for replay prioritization.

\begin{figure*}[!t]
\centering
\includegraphics[width=\linewidth]{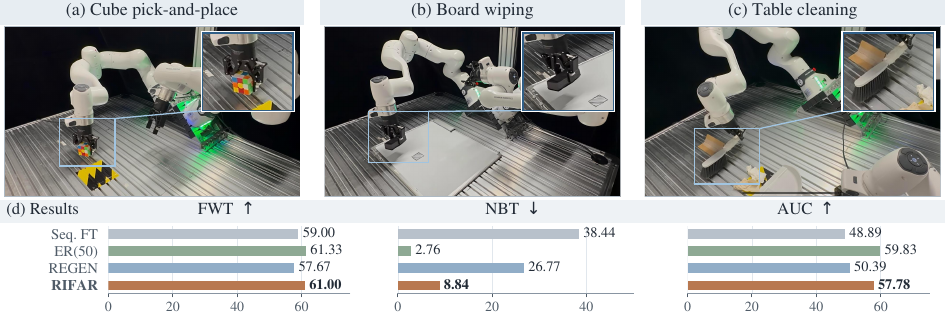}
\caption{Real-world continual learning on a Franka robot.
(a)--(c) Cube pick-and-place, board wiping, and table cleaning, with close-up insets.
(d) FWT, NBT, and AUC for sequential fine-tuning (Seq.\ FT), ER(50), REGEN, and RIFAR (Table~\ref{tab:realworld}), higher FWT/AUC and lower NBT are better.}
\label{fig:budget}
\end{figure*}

% \textbf{Reliability Analysis of IDM Screening.}
% In Figure~\ref{fig:idm_case}(a), generated observations suggest task completion, yet executing the paired actions in the simulator fails. This case motivates checking action--visual consistency when selecting generated trajectories. IDM screening retains a subset with a 76.9\% execution success rate, compared with 44.0\% for unfiltered candidates, a difference of 32.9 percentage points as shown in Figure~\ref{fig:idm_case}(b). The higher success rate of the screened subset supports IDM agreement as a reliability signal for generated replay.

\textbf{Reliability Analysis of IDM Screening.}
Figure~\ref{fig:idm_case}(a) shows a rollout that appears successful in the generated observations but fails when its actions are executed in the simulator. This mismatch illustrates why replay selection needs to account for action--visual consistency. As shown in Figure~\ref{fig:idm_case}(b), trajectories retained by IDM screening achieve a 76.9\% execution success rate, compared with 44.0\% for unfiltered candidates. This gain of 32.9 points suggests that agreement with the IDM helps identify generated trajectories whose actions are more likely to succeed during execution.

\textbf{Action-Drift Analysis.}
We compare fixed replay and drift-aware reselection on LIBERO-Goal under the same 800-step training budget. Both variants use current demonstrations and the same screened replay set for 500 steps. The control retains this set and the other reselects the ten trajectories per historical task with the highest $F_k(\widehat{\tau})$ from the full screened pool. Both then train for 300 steps with current data and their respective replay sets.

We evaluate both final policies against a shared reference using identical observations, proprioceptive states, and instructions from ten trajectories per evaluated task, with matched inference settings. Following Eq.~\eqref{eq:drift}, each input's drift is the normalized $\ell_2$ action difference averaged over the 16-step chunk. Mean and Max summarize the average and largest input-level drift. Tail averages the largest $20\%$, ranked separately for each variant. For fixed and reselected replay statistics $D_{\mathrm{fixed}}$ and $D_{\mathrm{reselect}}$, we define relative reduction for $D_{\mathrm{fixed}}>0$:
\begin{equation}
\textstyle
R_D(\%)=100\bigl(1-D_{\mathrm{reselect}}/D_{\mathrm{fixed}}\bigr).
\label{eq:drift_reduction}
\end{equation}
% Positive values indicate a reduction in action drift and negative values an increase. For task~3, Figure~\ref{fig:drift_reduction}(d) shows reductions in all three statistics across M6--M9. Some other tasks show increased maximum drift (Table~\ref{tab:detailed_drift}). Overall, the consistent drift reductions across M6--M9 indicate that drift-aware reselection better preserves historical behaviors, consistent with reduced forgetting and improved retention performance.
Positive values indicate reduced action drift on historical inputs, while negative values indicate an increase. Figure~\ref{fig:drift_reduction}(d) shows consistent reductions in all three drift statistics for task~3 across M6--M9. Table~\ref{tab:detailed_drift} also reports increased maximum drift for some other tasks. The drift analysis and retention results together suggest that drift-aware reselection helps preserve previously learned behaviors as the policy continues to learn new manipulation tasks.
\input{sections/realworld}

%% file: tables/table1.tex
\begin{table*}[!t]
\centering

\RIFARTable
\begin{tabularx}{\linewidth}{l*{9}{C}}
\toprule
& \multicolumn{3}{c}{\textbf{LIBERO-Goal}} & \multicolumn{3}{c}{\textbf{LIBERO-Object}} & \multicolumn{3}{c}{\textbf{LIBERO-Spatial}} \\
\cmidrule(lr){2-4}\cmidrule(lr){5-7}\cmidrule(lr){8-10}
\rowcolor{TableHeader} \textbf{Method} & \textbf{FWT $\uparrow$} & \textbf{NBT $\downarrow$} & \textbf{AUC $\uparrow$} & \textbf{FWT $\uparrow$} & \textbf{NBT $\downarrow$} & \textbf{AUC $\uparrow$} & \textbf{FWT $\uparrow$} & \textbf{NBT $\downarrow$} & \textbf{AUC $\uparrow$} \\
\midrule
Seq. FT & 90.4 & 99.3 & 40.0 & 98.73 & 83.65 & 55.06 & \textbf{94.67} & 97.88 & 44.28 \\
EWC & 94.4 & 97.46 & 43.77 & 98.33 & 82.95 & 55.07 & 93.87 & 97.9 & 44.07 \\
PackNet & 76.4 & 88.87 & 31.75 & 98.73 & 81.63 & 56.04 & 81.07 & 93.91 & 37.44 \\
\addlinespace[2pt]
ER(3) & 94.8 & 15.81 & 85.74 & 98.73 & 9.51 & 93.44 & 88.27 & 10.66 & 83.13 \\
ER(10) & 93.6 & 4.92 & 90.70 & 98.73 & 2.21 & 97.52 & 92.67 & 3.52 & \textbf{90.81} \\
ER(50) & 94.4 & \textbf{4.62} & \textbf{91.8} & \textbf{99.13} & \textbf{-0.49} & \textbf{99.35} & 90.67 & \textbf{-0.28}  & 90.22 \\
\addlinespace[2pt]
REGEN & 93.6 & 46.7 & 67.8 & 98.33 & 10.67 & 92.44 & 89.47 & 20.38 & 78.69 \\
\midrule
\rowcolor{RifarRow} \textcolor{RifarInk}{\textbf{RIFAR}} & \textbf{95.2} & 7.45 & 90.97 & \textbf{99.13} & 2.08 & 97.83 & 92.67 & 7.96 & 88.62 \\
\bottomrule
\end{tabularx}
\caption{
Continual learning performance on LIBERO-Goal, LIBERO-Object, and LIBERO-Spatial, evaluated using FWT, NBT, and AUC.
ER($n$) denotes rehearsal with $n$ complete demonstrations per historical task.
Results are averaged over three random seeds, and bold values indicate the best result in each column.
}
\label{tab:main}
\end{table*}

%% file: tables/table2.tex
\begin{table*}[!t]
\centering

\RIFARTable

\begin{tabularx}{\linewidth}{lCCC CCC CCC}
\toprule
&
\multicolumn{3}{c}{\textbf{LIBERO-Goal}} &
\multicolumn{3}{c}{\textbf{LIBERO-Object}} &
\multicolumn{3}{c}{\textbf{LIBERO-Spatial}} \\
\cmidrule(lr){2-4}
\cmidrule(lr){5-7}
\cmidrule(lr){8-10}
\rowcolor{TableHeader}
\textbf{Method} &
\textbf{FWT$\uparrow$} & \textbf{NBT$\downarrow$} & \textbf{AUC$\uparrow$}
&
\textbf{FWT$\uparrow$} & \textbf{NBT$\downarrow$} & \textbf{AUC$\uparrow$}
&
\textbf{FWT$\uparrow$} & \textbf{NBT$\downarrow$} & \textbf{AUC$\uparrow$}
\\
\midrule

Baseline (Anchor only)
& 93.0 & 43.5 & 69.5
& 98.33 & 8.79 & 93.27
& 90.01 & 18.37 & 80.36
\\

w/ IDM
& 93.2 & 10.61 & 87.14
& 98.73 & 6.95 & 94.85
& 91.07 & 11.63 & 84.80
\\

w/ Drift priority
& 92.8 & 18.18 & 82.49
& \textbf{99.53} & 4.14 & 97.06
& 87.07 & 12.81 & 80.59
\\

\addlinespace[2pt]
\rowcolor{RifarRow}
\textcolor{RifarInk}{\textbf{RIFAR (full)}}
& \textbf{95.2} & \textbf{7.45} & \textbf{90.97}
& 99.13 & \textbf{2.08} & \textbf{97.83}
& \textbf{92.67} & \textbf{7.96} & \textbf{88.62}
\\

\bottomrule
\end{tabularx}
\caption{Component ablation on three LIBERO suites. The baseline uses only
20 initial-state anchors for replay without IDM screening or drift-aware
selection. Bold indicates the best reported
value for each metric.}
\label{tab:ablation}
\end{table*}

%% file: tables/table8.tex
\begin{table}[!b]
\centering

\begingroup
\RIFARTable
\renewcommand{\arraystretch}{1.02}
\setlength{\tabcolsep}{1.8pt}
\begin{tabular*}{\columnwidth}{@{\extracolsep{\fill}}c
>{\columncolor{IDMFill}\centering\arraybackslash}c
cc
>{\columncolor{DriftFill}\centering\arraybackslash}c
cc@{}}
\toprule
\multirow{2}{*}{\textbf{Stage}}
& \multicolumn{3}{c}{\cellcolor{IDMFill}\textcolor{SuiteObjectInk}{\textbf{IDM reliability}}}
& \multicolumn{3}{c}{\cellcolor{DriftFill}\textcolor{SuiteGoalInk}{\textbf{Drift importance}}} \\
\cmidrule(lr){2-4}
\cmidrule(lr){5-7}
& \textbf{High} & \textbf{Random} & \textbf{Low}
& \textbf{High} & \textbf{Random} & \textbf{Low} \\
\midrule
G6 & 0.894 & 0.874 & 0.854 & 0.940 & 0.909 & 0.874 \\
O6 & 0.973 & 0.929 & 0.914 & 0.969 & 0.951 & 0.949 \\
S6 & 0.920 & 0.917 & 0.866 & 0.951 & 0.914 & 0.911 \\
\bottomrule
\end{tabular*}
\endgroup
\caption{Replay-selection success rates at stage 1 on LIBERO-Goal (G6), Object (O6), and Spatial (S6).}
\label{tab:selection_validation}
\end{table}

%% file: sections/realworld.tex
\subsection{Real-World Continual Learning}
\label{sec:realworld}

\input{tables/table_realworld}

We evaluate Cosmos Policy on three real-world tasks using a Franka robot (Figure~\ref{fig:budget}). Cube pick-and-place requires precise grasping and placement; board wiping requires controlled motion while maintaining eraser contact with the board; and table cleaning involves sweeping with a broom. Their differing motion and contact requirements test whether a shared policy can acquire new behaviors while retaining earlier skills. We compare RIFAR with sequential fine-tuning, ER(50), and REGEN. ER(50) replays 50 complete demonstrations per historical task, whereas REGEN and RIFAR replay generated historical experience.

After each stage, we evaluate all methods on the tasks seen so far and report FWT, NBT, and AUC. Table~\ref{tab:realworld} shows that REGEN reduces forgetting relative to sequential fine-tuning, while RIFAR improves both the acquisition of new tasks and the retention of earlier skills. Compared with REGEN, RIFAR increases AUC from 50.39 to 57.78 and lowers NBT from 26.77 to 8.84. FWT also improves from 57.67 to 61.00, indicating better performance on newly introduced tasks. ER(50) performs best overall, with an AUC of 59.83. RIFAR comes within 2.05 AUC points while retaining only compact historical contexts. It offers a practical alternative to storing complete demonstrations, reducing the amount of historical data needed to maintain performance across successive tasks.

% After each stage, we evaluate all methods on the tasks seen so far and report FWT, NBT, and AUC. Table~\ref{tab:realworld} shows that REGEN reduces forgetting relative to sequential fine-tuning, while RIFAR improves both the acquisition of new tasks and the retention of earlier skills. Compared with REGEN, RIFAR increases AUC from 50.39 to 57.78 and lowers NBT from 26.77 to 8.84. FWT also improves from 57.67 to 61.00, indicating better performance on newly introduced tasks. ER(50) performs best overall, with an AUC of 59.83. RIFAR comes within 2.05 AUC points while retaining only compact historical contexts. It therefore offers a practical alternative to storing complete demonstrations, reducing the amount of historical data needed to maintain performance across successive tasks.

%% file: tables/table_realworld.tex
\begin{table}[!t]
\centering

\RIFARTable
\setlength{\tabcolsep}{3pt}
\begin{tabularx}{\columnwidth}{@{}lCCC@{}}
\toprule
\rowcolor{TableHeader}
\textbf{Method}
& \textbf{FWT $\uparrow$}
& \textbf{NBT $\downarrow$}
& \textbf{AUC $\uparrow$} \\
\midrule
Sequential fine-tuning
& 59.00
& 38.44
& 48.89 \\
ER(50)
& \textbf{61.33}
& \textbf{2.76}
& \textbf{59.83} \\
REGEN
& 57.67
& 26.77
& 50.39 \\
\midrule
\rowcolor{RifarRow}
\textcolor{RifarInk}{\textbf{RIFAR (Ours)}}
& 61.00
& 8.84
& 57.78 \\
\bottomrule
\end{tabularx}
\caption{Real-world continual learning on three manipulation tasks using Cosmos Policy on a Franka robot. Bold indicates the best result for each metric.}
\label{tab:realworld}
\end{table}

%% file: sections/conclusion.tex
\section{Conclusion}

RIFAR uses the generative capacity of WAMs to preserve earlier manipulation skills while learning new tasks. It reconstructs past-task trajectories from compact historical contexts and uses an inverse-dynamics model to screen them for action--visual consistency. Action drift then guides replay toward earlier behaviors most affected by new-task learning. Across three LIBERO suites and real-world Franka experiments, RIFAR achieves higher AUC and lower NBT than REGEN while remaining competitive with real-data replay. On LIBERO-Goal, it reaches 90.97 AUC while retaining only 320 historical time steps per task, just 4.9\% of those retained by ER(50). This exceeds the 90.70 AUC of ER(10), which retains approximately 1,300 steps per task. These findings support a broader vision of genuine embodied agency built on continual learning that transforms experience into lasting capabilities across evolving tasks and environments.

%% file: sections/limitations.tex
\section*{Limitations}

Although RIFAR improves continual learning performance, several limitations
remain. First, IDM-based reliability screening can improve the quality of
generated replay trajectories, but it cannot completely remove unreliable
samples. As continual learning proceeds, the IDM is updated using both newly
collected demonstrations and WAM-generated replay data. Since generated
trajectories may be less reliable than real demonstrations, they can introduce
noise into IDM training and affect subsequent reliability estimation. 

Second, the current world action model relies on value-based termination during
autoregressive generation. The predicted value may
become high before the task is actually completed, causing premature termination.

Future
work could explore stronger trajectory verification and reliability models, while introducing explicit task-success verification to improve termination reliability and trajectory completeness.

%% file: sections/appendix_lt_A.tex
\appendix
\setlength{\parindent}{0pt}
\section{Implementation Specification}
\label{app:implementation}

\subsection{LIBERO Benchmark}
\label{app:libero}

LIBERO is a language-conditioned robot manipulation benchmark designed to
evaluate lifelong robot learning under sequential task adaptation. It contains
multiple task suites with different sources of task variation. In this work, we
evaluate RIFAR on three LIBERO suites: LIBERO-Goal, LIBERO-Object, and
LIBERO-Spatial.

LIBERO-Goal focuses on goal-level manipulation variations, where the robot
needs to learn diverse manipulation objectives, such as object placement,
drawer operation, and object interaction. LIBERO-Object evaluates object-level
generalization by requiring similar manipulation skills to be transferred across
different object categories. LIBERO-Spatial emphasizes spatial reasoning, where
the robot needs to perform manipulation under different object locations and
spatial configurations.

Each suite contains 10 language-conditioned manipulation tasks. Following our
continual learning protocol, Tasks 0--5 are jointly used for initial policy
training, while Tasks 6--9 are sequentially introduced during continual
learning. The complete task instructions used in our experiments are listed in
Table~\ref{tab:libero_tasks}.

\subsection{Real-World Experimental Details}
\label{app:real_world_details}

We conduct real-world continual learning experiments on a Franka robot using
the same Cosmos Policy architecture and 16-step action horizon as in simulation.
The three tasks are learned sequentially in the order of cube pick-and-place,
board wiping, and table cleaning, with 50 demonstrations per task. Each task is evaluated over 50 trials. Initial states are manually reset and
therefore contain small variations from the training configurations. After each
learning stage, all previously seen tasks are evaluated, and task success is
determined manually using the same criteria for all methods.

%% file: sections/appendix_lt_B.tex
\section{Training Details}
\label{app:training}

\subsection{Training Protocol}
\label{app:training_protocol}

Before continual learning, we first train a base policy on the initial task set
of each LIBERO suite. The policy is initialized from the pretrained
Cosmos-Predict2-2B-Video2World checkpoint and fine-tuned using the
Cosmos-Policy framework. For each LIBERO suite, Tasks 0--5 are jointly used for
base policy training, providing the initial manipulation capabilities for
subsequent continual learning experiments. The training data consists of rollout trajectories. For each task, 50 successful trajectories are used during
training. Demonstrations and rollout trajectories are sampled with equal
probability during optimization. The policy predicts action chunks with a
horizon of $H=16$.

After obtaining the base policy checkpoint, we perform sequential continual
learning on Tasks 6--9. At each continual learning stage, the model is trained
using newly introduced task demonstrations together with historical replay
trajectories. Each continual learning stage consists of two optimization periods separated by
a replay update step, with a total of 800 optimization steps. The first
optimization period contains 500 steps. During the first 300 steps, the ratio
between current-task data and replay data is set to 6:4. For the remaining 200
steps, the ratio is changed to 5:5. After this period, the historical replay
set is updated using drift-aware replay selection. The model is then further
optimized for 300 steps using the updated replay set with a current-task to
replay-data ratio of 5:5.

\subsection{Optimization Hyperparameters}
\label{app:hyperparameters}

The main optimization configurations used in base policy training and
continual learning are summarized in Table~\ref{tab:training_hyperparams}. The
same configuration is applied across all LIBERO suites.
\input{tables/table13}
\input{tables/table15}

\subsection{IDM Training Procedure}
\label{app:idm_training}

The inverse dynamics model (IDM) is trained to evaluate the reliability of
generated replay trajectories. Initially, the IDM is trained using real
demonstrations from LIBERO Tasks 0--5. Given two image observations separated
by 16 time steps, the IDM predicts the corresponding intermediate action
sequence, where each action is 7-dimensional.

During continual learning, the IDM is updated sequentially at  each learning
stage. For each new task, the update set combines 50 real
demonstrations from the current task with replay trajectories selected from
previous tasks. The top-10 trajectories from each previous task are selected
according to the IDM reliability scores and added to the update set. After
training on this dataset, the updated IDM checkpoint is used for reliability
evaluation in subsequent replay selection stages. For real-world experiments, the IDM is initially trained on Task 1 and then continually updated as Tasks 2 and 3 are introduced following the same procedure.

The IDM architecture and optimization hyperparameters are summarized in
Table~\ref{tab:idm_hyperparams}.
\input{tables/table_idm}
\begin{figure*}[!tb]
\centering

\includegraphics[width=\linewidth]{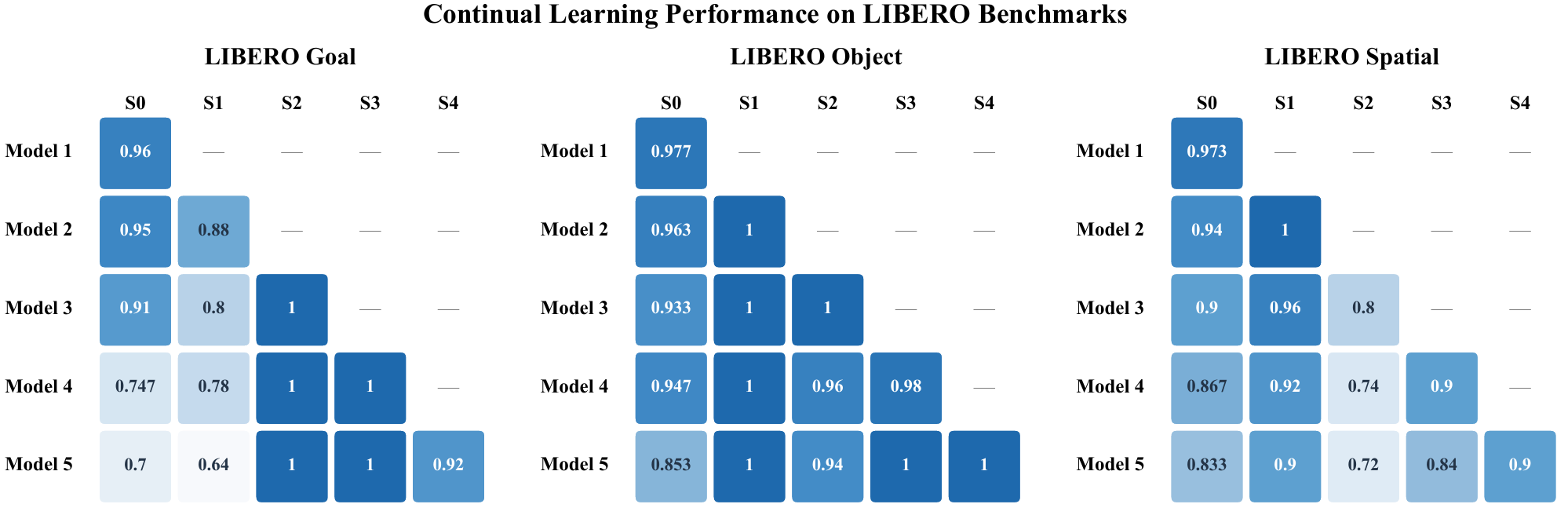}

\caption{
Continual learning performance on LIBERO-Goal, LIBERO-Object, and LIBERO-Spatial.
Rows represent model checkpoints after successive learning stages, and columns denote task groups S0--S4.
Each cell reports the corresponding task success rate, while not-yet-introduced tasks are masked.
The lower-triangular matrices visualize both new-task acquisition and retention of previously learned tasks.
Results are averaged over three random seeds.
}
\label{fig:task_success_matrix}

\end{figure*}
\subsection{Complete RIFAR Algorithm}
\label{sec:rifar_algorithm}

Algorithm~\ref{alg:rifar} summarizes the complete RIFAR procedure.
RIFAR first reconstructs historical trajectories from compact prefixes
and filters them according to inverse-dynamics consistency.
After reliability-aware adaptation, it measures the behavioral drift
induced by new-task learning and prioritizes high-drift historical
trajectories for the second replay phase.

\input{sections/algorithm}
\subsection{Evaluation Protocol}
\label{app:evaluation}

We evaluate continual learning performance following the sequential task
adaptation protocol described in the main paper. For each LIBERO suite, the
model is first trained on the initial task set (Tasks 0--5) and evaluated to
obtain the initial performance. Then, new tasks (Tasks 6--9) are introduced
sequentially, and the model is evaluated after each task adaptation stage.
During evaluation, the model is tested on all previously encountered tasks,
including both the newly acquired task and historical tasks. Therefore, each
evaluation stage measures both the acquisition ability on the current task and
the retention of previously learned manipulation skills. All compared methods
follow the same task order, evaluation environments, language instructions, and
observation modalities to ensure a fair comparison.

For continual learning evaluation, we report three complementary metrics:
forward transfer (FWT), normalized backward transfer (NBT), and area under the
continual learning curve (AUC). FWT measures the ability of a model to acquire
new tasks after continual updates. NBT evaluates the performance degradation of
previously learned tasks after learning new tasks. AUC summarizes the overall
continual learning performance throughout the sequential adaptation process by
considering both task acquisition and knowledge retention. Higher FWT and AUC
values indicate better continual learning performance, while lower NBT values
indicate less degradation.

The corresponding formulations are given below. Let $r_{n,n}$ denote the
performance on stage $n$ after learning stage $n$, and $r_{p,n}$ denote the
performance on stage $p$ after learning stage $n$. FWT is computed as
\begin{equation}
FWT=\frac{1}{N}\sum_{n=1}^{N}r_{n,n}.
\label{eq:fwt}
\end{equation}

NBT measures the normalized performance degradation of previous stages:
\begin{equation}
NBT_n=\frac{1}{N-n}\sum_{p=n+1}^{N}
\frac{r_{n,n}-r_{p,n}}{r_{n,n}}.
\label{eq:nbt_stage}
\end{equation}

The overall NBT is averaged over all previous stages:
\begin{equation}
NBT=\frac{1}{N-1}\sum_{n=1}^{N-1}NBT_n.
\label{eq:nbt}
\end{equation}
Finally, AUC summarizes the overall continual learning performance:

\begin{equation}
AUC=\frac{1}{N}\sum_{n=1}^{N}
\frac{r_{n,n}+\sum_{p=n+1}^{N}r_{p,n}}
{N-n+1}.
\label{eq:auc}
\end{equation}

Here, $N$ denotes the total number of sequential learning stages, and
$r_{p,n}$ represents the performance of stage $p$ after the model has been
trained up to stage $n$.

In addition to aggregate metrics, we report per-task success rates to provide a
more detailed analysis of task-wise performance.

%% file: tables/table13.tex
\begin{table}[t]
\centering
\small

\RIFARTable

% =========================
% Adjustable parameters
% =========================
\setlength{\tabcolsep}{3pt}
\renewcommand{\arraystretch}{1.15}
% =========================

\begin{tabularx}{\columnwidth}{
>{\raggedright\arraybackslash}p{0.22\columnwidth}
>{\raggedright\arraybackslash}p{0.35\columnwidth}
>{\raggedright\arraybackslash}X
}

\toprule
\rowcolor{TableHeader}

\textbf{Category}
&
\textbf{Hyperparameter}
&
\textbf{Value}
\\

\midrule

\multirow{3}{*}[-1.2ex]{\raggedright \textcolor{TableMuted}{Model}}
& Backbone
& Cosmos-Predict2-2B-Video2World
\\

& Framework
& Cosmos-Policy
\\

& Action horizon
& 16
\\

\midrule

\multirow{6}{*}[-1.0ex]{\raggedright \textcolor{SuiteObjectInk}{Base training}}
& GPUs
& 8
\\

& Batch size per GPU
& 15
\\

& Gradient accumulation
& 2
\\

& Effective batch size
& 240
\\

& Learning rate
& $1\times10^{-4}$
\\

& Training iterations
& 30K
\\

\midrule

\multirow{7}{*}[-2.0ex]{\raggedright
\textcolor{RifarInk}{\makecell[l]{Continual\\learning}}}
& GPUs
& 8
\\

& Batch size per GPU
& 15
\\

& Gradient accumulation
& 3
\\

& Total optimization steps/task
& 800
\\

& Warm-up steps
& 50
\\

& Peak learning rate
& $3.5\times10^{-5}$
\\

& Final learning rate
& $2.5\times10^{-5}$
\\

\bottomrule

\end{tabularx}

\caption{Optimization hyperparameters used in training.}
\label{tab:training_hyperparams}
\end{table}

%% file: tables/table15.tex
\begin{table*}[t]
\centering
\arrayrulecolor{TableRule}
\small

\renewcommand{\arraystretch}{1.05}

\setlength{\tabcolsep}{6pt}

\begin{tabularx}{\textwidth}{
>{\raggedright\arraybackslash}p{0.18\textwidth}
>{\centering\arraybackslash}p{0.08\textwidth}
>{\raggedright\arraybackslash}X
}

\toprule
\rowcolor{TableHeader}

\textbf{Suite}
&
\textbf{Task ID}
&
\textbf{Language Instruction}
\\

\midrule

\multirow{10}{*}{\textcolor{SuiteGoalInk}{LIBERO-Goal}}
&0&open the middle drawer of the cabinet\\
&1&put the bowl on the stove\\
&2&put the wine bottle on top of the cabinet\\
&3&open the top drawer and put the bowl inside\\
&4&put the bowl on top of the cabinet\\
&5&push the plate to the front of the stove\\
&6&put the cream cheese in the bowl\\
&7&turn on the stove\\
&8&put the bowl on the plate\\
&9&put the wine bottle on the rack\\

\midrule

\multirow{10}{*}{\textcolor{SuiteObjectInk}{LIBERO-Object}}
&0&pick up the alphabet soup and place it in the basket\\
&1&pick up the cream cheese and place it in the basket\\
&2&pick up the salad dressing and place it in the basket\\
&3&pick up the bbq sauce and place it in the basket\\
&4&pick up the ketchup and place it in the basket\\
&5&pick up the tomato sauce and place it in the basket\\
&6&pick up the butter and place it in the basket\\
&7&pick up the milk and place it in the basket\\
&8&pick up the chocolate pudding and place it in the basket\\
&9&pick up the orange juice and place it in the basket\\

\midrule

\multirow{10}{*}{\textcolor{SuiteSpatialInk}{LIBERO-Spatial}}
&0&pick up the black bowl between the plate and the ramekin and place it on the plate\\
&1&pick up the black bowl next to the ramekin and place it on the plate\\
&2&pick up the black bowl from table center and place it on the plate\\
&3&pick up the black bowl on the cookie box and place it on the plate\\
&4&pick up the black bowl in the top drawer of the wooden cabinet and place it on the plate\\
&5&pick up the black bowl on the ramekin and place it on the plate\\
&6&pick up the black bowl next to the cookie box and place it on the plate\\
&7&pick up the black bowl on the stove and place it on the plate\\
&8&pick up the black bowl next to the plate and place it on the plate\\
&9&pick up the black bowl on the wooden cabinet and place it on the plate\\

\bottomrule

\end{tabularx}

\caption{
Task instructions used in LIBERO continual learning experiments.
Tasks 0--5 are used for initial joint training, while Tasks 6--9 are
sequentially introduced during continual learning.
}
\label{tab:libero_tasks}
\end{table*}

%% file: tables/table_idm.tex
\begin{table}[t]
\centering
\small

\RIFARTable

\setlength{\tabcolsep}{5pt}
\renewcommand{\arraystretch}{1.05}

\begin{tabularx}{\linewidth}{
@{}
>{\raggedright\arraybackslash}p{0.52\linewidth}
>{\raggedright\arraybackslash}X
@{}
}

\toprule
\rowcolor{TableHeader}
\textbf{Hyperparameter} & \textbf{Value} \\
\midrule

Action Chunk Length
& 16 \\

Action / State Dimension
& 7 / 9 \\

Input Image Resolution
& $224\times224$ \\

Vision Backbone
& DINOv2 ViT-S/14 \\

Attention Heads
& 8 \\

Decoder Layers
& 2 \\

Dropout
& 0.3 \\

Global Batch Size
& 64 \\

GPU Configuration
& 8 GPUs \\

Optimizer
& AdamW \\

Initial Learning Rate
& $3\times10^{-4}$ \\

Weight Decay
& $1\times10^{-4}$ \\

Scheduler
& CosineAnnealingLR \\

Training Epochs
& 30 \\

Gradient Clipping
& 5.0 \\

\bottomrule

\end{tabularx}

\caption{IDM training hyperparameters.}
\label{tab:idm_hyperparams}

\end{table}

%% file: sections/algorithm.tex
\begin{algorithm}[t]
\caption{RIFAR: Reliability- and Drift-Aware Generative Replay}
\label{alg:rifar}
\small
\begin{algorithmic}[1]

\Require Current-task demonstrations $\mathcal{D}_k$,
         policy $\pi_{\theta_{k-1}}$,
         historical prefixes $\{\mathcal{P}_i\}_{i<k}$,
         inverse-dynamics model $f_{\mathrm{IDM}}^{k-1}$
\Ensure Updated policy $\pi_{\theta_k}$ and IDM $f_{\mathrm{IDM}}^k$

\State $\mathcal{R}_{\mathrm{rel}} \gets \emptyset$

\For{each previous task $i<k$}
    \State $\mathcal{C}_i \gets
    \textsc{Generate}(
    \pi_{\theta_{k-1}}, \mathcal{P}_i)$

    \State $s_{\mathrm{rel}}(\tau) \gets
    \textsc{IDMScore}(f_{\mathrm{IDM}}^{k-1},\tau)$,
    $\forall \tau \in \mathcal{C}_i$

    \State $\mathcal{R}_{\mathrm{rel}}^i \gets
    \textsc{SelectReliable}
    (\mathcal{C}_i,s_{\mathrm{rel}})$

    \State $\mathcal{R}_{\mathrm{rel}} \gets
    \mathcal{R}_{\mathrm{rel}}
    \cup \mathcal{R}_{\mathrm{rel}}^i$
\EndFor

\State $f_{\mathrm{IDM}}^k \gets
\textsc{TrainIDM}(
f_{\mathrm{IDM}}^{k-1},
\mathcal{D}_k \cup \mathcal{R}_{\mathrm{rel}})$

\Statex
\Statex \textbf{Phase I: Reliability-aware adaptation}

\State $\theta^{\mathrm{pre}} \gets \theta_{k-1}$

\State $\theta^{\mathrm{mid}} \gets
\textsc{Train}(
\theta^{\mathrm{pre}},
\mathcal{D}_k \cup \mathcal{R}_{\mathrm{rel}},
500)$

\Statex
\Statex \textbf{Phase II: Drift-aware replay selection}

\State $\mathcal{R}_{\mathrm{drift}} \gets \emptyset$

\For{each previous task $i<k$}
    \For{each $\tau \in \mathcal{R}_{\mathrm{rel}}^i$}
        \State $s_{\mathrm{drift}}(\tau) \gets
        \textsc{DriftScore}
        (\pi_{\theta^{\mathrm{pre}}},
         \pi_{\theta^{\mathrm{mid}}},
         \tau)$
    \EndFor

    \State $\mathcal{R}_{\mathrm{drift}}^i \gets
    \textsc{SelectHighDrift}
    (\mathcal{R}_{\mathrm{rel}}^i,
     s_{\mathrm{drift}})$

    \State $\mathcal{R}_{\mathrm{drift}} \gets
    \mathcal{R}_{\mathrm{drift}}
    \cup \mathcal{R}_{\mathrm{drift}}^i$
\EndFor

\State $\theta_k \gets
\textsc{Train}(
\theta^{\mathrm{mid}},
\mathcal{D}_k \cup \mathcal{R}_{\mathrm{drift}},
300)$

\State \Return $\pi_{\theta_k}, f_{\mathrm{IDM}}^k$

\end{algorithmic}
\end{algorithm}

%% file: sections/appendix_lt_C.tex
\section{Additional Results}

\subsection{Task-wise Continual Learning Performance}
\label{app:success}

To provide a detailed view of continual learning behavior, we report the
task-wise success rate matrices on the three LIBERO suites. Each matrix records
the success rates evaluated after each continual learning stage. Rows correspond
to evaluation stages, while columns correspond to previously encountered tasks.
Therefore, the matrices reflect both the acquisition of newly introduced tasks
and the retention of historical manipulation skills.

\begin{figure*}[!tb]
\centering

\includegraphics[width=\linewidth]{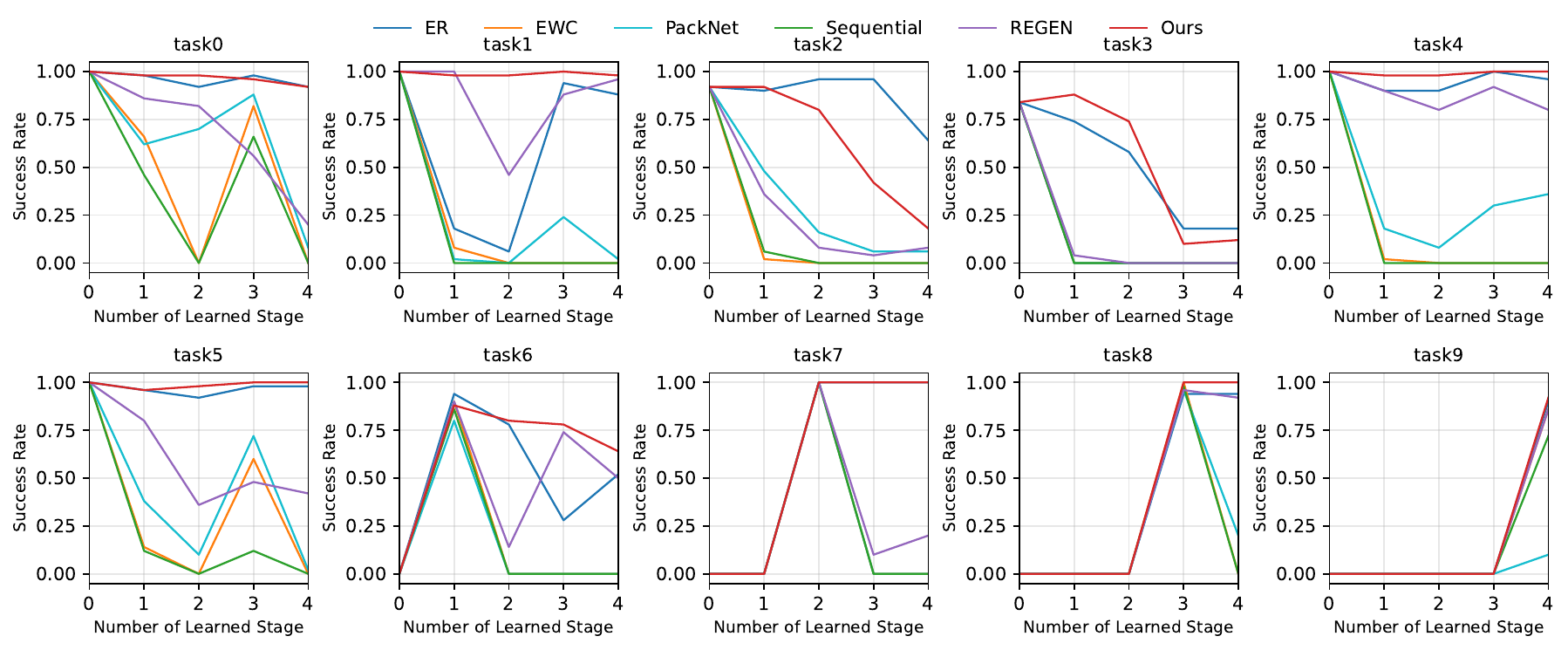}

\caption{
Per-task success rate comparison on LIBERO-Goal. ER denotes ER(3).
}
\label{fig:goal}

\end{figure*}

\begin{figure*}[!tb]
\centering

\includegraphics[width=\linewidth]{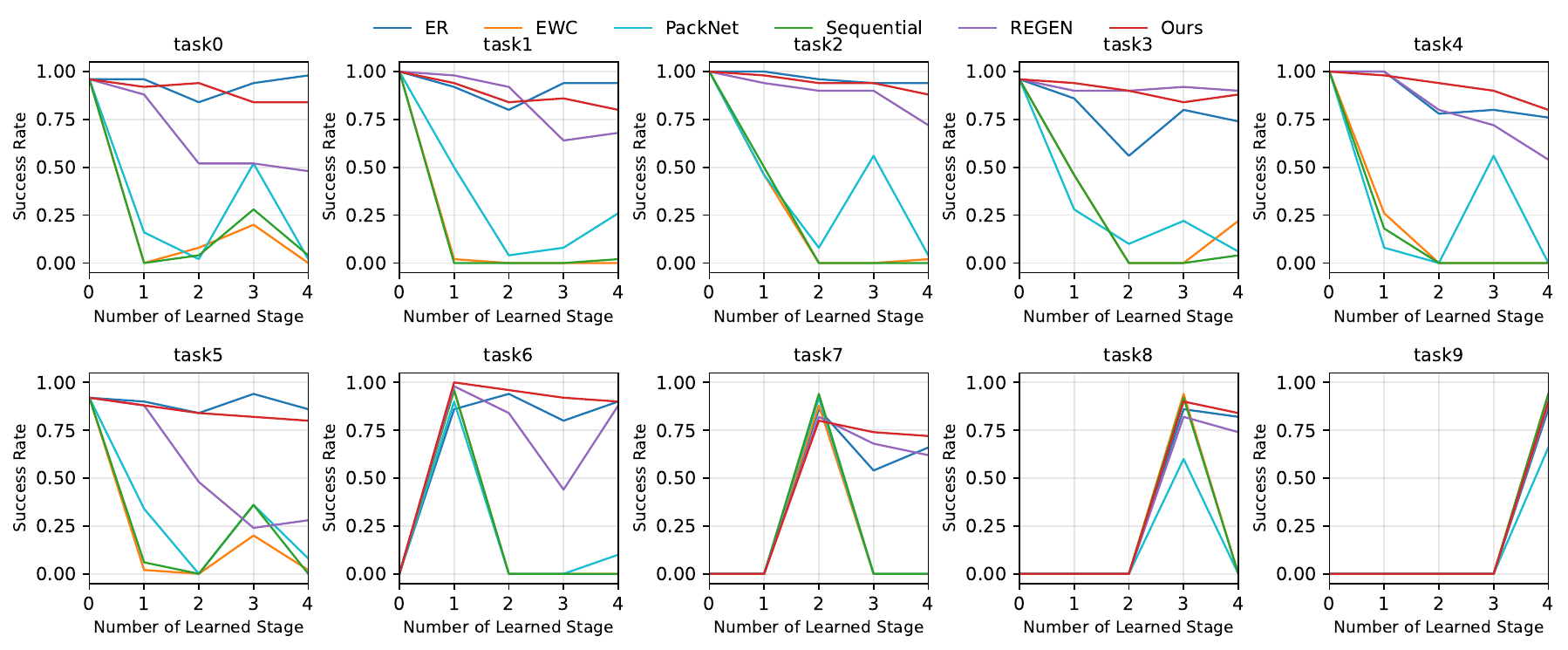}

\caption{
Per-task success rate comparison on LIBERO-Spatial. ER denotes ER(3).
}
\label{fig:spatial}

\end{figure*}

For compact visualization, we represent the performance evolution as $5\times5$
matrices. As shown in Figure~\ref{fig:task_success_matrix}, the first row corresponds to the initial policy after joint training
on Tasks 0--5, and the following rows correspond to sequential adaptation after
learning new tasks (Tasks 6--9). All success rates are averaged over three
random seeds. The same evaluation protocol is applied to LIBERO-Goal,
LIBERO-Object, and LIBERO-Spatial. The task-wise success rates of the individual components are
further reported in Figures~\ref{fig:goal}, \ref{fig:object}, and
\ref{fig:spatial} for LIBERO-Goal, LIBERO-Object, and LIBERO-Spatial,
respectively.

\subsection{WAM Rollout and Termination Criteria}
\label{app:wam_rollout}

For each task, we generate trajectories using the World Action Model
(WAM) with a maximum rollout horizon of \(H=200\) environment steps.
At each step \(t\), the WAM produces a value prediction \(v_t\) after
executing the predicted action. The same termination rule is applied to
all tasks and random seeds. For compactness, we define the predicate
\begin{equation}
\mathcal{H}_{t}^{(k)}
\triangleq
\bigwedge_{i=0}^{k-1}
\left(v_{t+i}>0.99\right),
\end{equation}
which indicates that the value prediction remains greater than \(0.99\)
for \(k\) consecutive steps. A rollout is considered successful if one of
the following two conditions is satisfied:
\begin{equation}
\begin{aligned}
\mathrm{Success}\iff{}&
\left(\exists t:\,
v_t\geq 1.0
\land \mathcal{H}_{t+1}^{(3)}\right)\\
&{}\lor
\left(\exists t:\,
\mathcal{H}_{t}^{(6)}\right).
\end{aligned}
\end{equation}

The first condition requires the value prediction to reach at least
\(1.0\), followed by three consecutive predictions greater than
\(0.99\). The second condition accepts six consecutive value predictions
greater than \(0.99\), without requiring an exact value of \(1.0\). A rollout satisfying either condition is labeled as
successful. If neither condition is satisfied within \(H=200\) steps, the
rollout is terminated due to timeout and assigned the label
\texttt{false}. Timeout rollouts are not counted as successful generated
trajectories and are assigned the \texttt{false} label before subsequent IDM-based reliability filtering. Generation success rates are reported in Table~\ref{tab:wam_generation_success}.
\input{tables/1234}
\subsection{IDM-based Trajectory Reliability Scoring}
\label{app:idm_scoring}

We use an IDM-based scoring model to evaluate replay trajectory reliability.
The model predicts 16-step action chunks with an action dimension of 7.
\input{tables/table9}
For each chunk, the normalized action prediction error is
\begin{equation}
\textstyle
E_{\mathrm{IDM}}=d^{-1/2}\operatorname{mean}_{t}\|\hat a_t-a_t\|_2,
\label{eq:idm_appendix_error}
\end{equation}
where $\hat a_t$ and $a_t$ are the normalized predicted and executed action vectors. This is the root-mean-square error across action dimensions, averaged over time. The checkpoint-calibrated reliability score is
\begin{equation}
\textstyle
Q_{\mathrm{IDM}}=\sigma((b-sE_{\mathrm{IDM}})/T),
\label{eq:idm_appendix_score}
\end{equation}
with $s=8.6403$, $b=6.5952$, and $T=1.0$ are fixed empirical parameters. These values specify the same sigmoid mapping as Eq.~\eqref{eq:idm_score}, with $\alpha=b/T$ and $\beta=s/T$. Lower prediction errors therefore produce higher IDM scores.

For trajectory-level scoring, we first smooth the scores using a three-chunk trailing moving average. For $N$ chunk scores, let $Q^{\mathrm{sm}}_{(j)}$ denote the $j$-th smallest smoothed value and set $k=\max(1,\lceil0.2N\rceil)$. The trajectory soft score is the mean of the lowest $20\%$:
\begin{equation}
\textstyle
Q_{\mathrm{soft}}=k^{-1}\sum\nolimits_{j=1}^{k}Q^{\mathrm{sm}}_{(j)}.
\label{eq:idm_soft}
\end{equation}
Here, $Q_{\mathrm{soft}}$ corresponds to \texttt{soft\_score} in the implementation.
This design emphasizes locally unreliable segments instead of averaging out
short-term failures. 

A trajectory is classified as reliable when
$Q_{\mathrm{soft}}\geq0.65$ and the longest consecutive sequence with
$Q_{\mathrm{IDM}}<0.40$ is shorter than 3 chunks.
For drift evaluation, chunks with $Q_{\mathrm{IDM}}<0.60$ are regarded as
low-score drift chunks, and the allowed number is
$\max(1,\lfloor0.2N_{\mathrm{judged}}\rfloor)$.
Consecutive drift length is only recorded for analysis and is not used as an
additional rejection condition.

If the strict reliable condition is not satisfied, trajectories with
$Q_{\mathrm{soft}}\geq0.40$, fewer than four consecutive low-score chunks, and a
passed drift gate are categorized as suspect; otherwise they are classified as
unreliable. The number of trajectories satisfying the reliable criterion after
IDM-based filtering for each continual learning setting is reported in
Table~\ref{tab:idm_reliable_ratio}. When fewer than 10 reliable trajectories are selected for a task, the selected
trajectories are sequentially duplicated until the replay set reaches the
required size of 10 trajectories.
Only when no
reliable trajectories are available for a task do we use suspect trajectories
as a fallback to construct the replay set.
\subsection{Drift Analysis across Continual Learning Stages}
\label{app:drift_analysis}

To further analyze the effect of drift-aware replay, we measure the behavioral
drift of different continual learning models with and without drift-aware
selection. D$-$ denotes the replay strategy without drift-aware selection,
while D$+$ denotes the proposed drift-aware replay strategy. For Task0--Task5,
the drift is calculated between each continual learning model and the
corresponding base model. For Task6 and later stages, the drift is measured
between consecutive continual learning models, reflecting the additional
behavioral changes introduced by subsequent task learning.

As summarized in Table~\ref{tab:detailed_drift}, drift-aware replay generally reduces behavioral drift relative to the old-model baseline. While several individual stages exhibit small increases in specific statistics, the overall Mean, Max, and Tail drift remain lower with drift-aware replay. This suggests that explicitly prioritizing high-drift trajectories directs replay toward historical behaviors that are more affected by new-task adaptation, thereby improving behavioral retention. 

\subsection{Replay Selection Overlap Analysis}
\label{app:replay_overlap}

To analyze the consistency between different replay selection stages, we report
the number of overlapping trajectories between replay1 and replay2 after the
final replay selection process. Each entry in Table~\ref{tab:replay_overlap}
denotes the number of shared trajectories between the two replay sets.

The overlap statistics characterize the consistency of the replay selection
pipeline. The shared\input{tables/table10} trajectories indicate samples consistently considered
valuable by both selection stages, while the remaining trajectories reflect the
additional filtering effect introduced by subsequent selection criteria.
\subsection{Computational Resources}
We use 16 PRO 5000 GPUs: 8 for policy training and 8 for replay generation and evaluation. All methods use identical optimization settings. Rollout generation, IDM updating, reliability screening, and drift scoring take substantially less time than policy training. Generated replay is discarded after each adaptation stage.
\input{tables/table12}

%% file: tables/1234.tex
\begin{table}[t]
\centering
\small

\RIFARTable

\setlength{\tabcolsep}{4pt}
\renewcommand{\arraystretch}{1.08}

\begin{tabularx}{\linewidth}{
    @{}
    >{\raggedright\arraybackslash}X
    >{\centering\arraybackslash}X
    >{\centering\arraybackslash}X
    >{\centering\arraybackslash}X
    >{\centering\arraybackslash}X
    @{}
}
\toprule
\rowcolor{TableHeader}
\textbf{Suite} &
\textbf{Model6} &
\textbf{Model7} &
\textbf{Model8} &
\textbf{Model9} \\
\midrule

Goal
& 93.3
& 86.8
& 76.6
& 82.2 \\

Object
& 86.7
& 71.1
& 65.0
& 57.5 \\

Spatial
& 89.2
& 85.0
& 76.3
& 71.9 \\

\bottomrule
\end{tabularx}

\caption{Value-based termination rate (\%) of WAM recursive rollouts across continual learning stages. Each historical task is given
40 generation attempts.}
\label{tab:wam_generation_success}
\end{table}

%% file: tables/table9.tex
\begin{table*}[t]
\centering
\arrayrulecolor{TableRule}
\small

\renewcommand{\arraystretch}{1.1}
% 调整这里控制列间距
\setlength{\tabcolsep}{6.5pt}
\resizebox{\textwidth}{!}{%
\begin{tabular}{llccccccccc}
\toprule
\rowcolor{TableHeader}
\textbf{Data}
&
\textbf{Setting}
&
\textbf{Task0}
&
\textbf{Task1}
&
\textbf{Task2}
&
\textbf{Task3}
&
\textbf{Task4}
&
\textbf{Task5}
&
\textbf{Task6}
&
\textbf{Task7}
&
\textbf{Task8}
\\
\midrule
\multirow{4}{*}{\textcolor{SuiteGoalInk}{Goal}}
& Model6
&36/37
&40/40
&40/40
&34/34
&40/40
&30/33
&--
&--
&--
\\
& Model7
&36/36
&40/40
&33/34
&24/24
&40/40
&28/31
&38/38
&--
&--
\\
& Model8
&25/31
&40/40
&35/35
&13/13
&37/37
&6/15
&35/35
&39/39
&--
\\
& Model9
&14/24
&40/40
&33/35
&33/33
&33/34
&2/20
&30/33
&36/37
&40/40
\\
\midrule
\multirow{4}{*}{\textcolor{SuiteObjectInk}{Object}}
& Model6
&30/33
&34/36
&34/35
&35/36
&34/35
&30/33
&--
&--
&--
\\
& Model7
&21/26
&27/29
&28/30
&30/32
&24/29
&21/27
&23/26
&--
&--
\\
& Model8
&19/24
&22/26
&20/24
&27/30
&25/30
&3/20
&10/22
&30/32
&--
\\
& Model9
&7/22
&4/20
&10/23
&21/24
&4/20
&2/21
&3/20
&23/28
&25/29
\\
\midrule
\multirow{4}{*}{\textcolor{SuiteSpatialInk}{Spatial}}
& Model6
&35/36
&39/39
&38/38
&40/40
&22/28
&32/33
&--
&--
&--
\\
& Model7
&35/35
&35/35
&30/31
&37/37
&23/28
&37/37
&34/35
&--
&--
\\
& Model8
&22/28
&34/34
&33/34
&28/30
&19/26
&16/24
&35/35
&32/33
&--
\\
& Model9
&24/30
&37/37
&19/25
&21/27
&17/28
&5/22
&25/26
&33/33
&30/31
\\
\bottomrule
\end{tabular}%
}
\caption{
Reliable trajectory counts after IDM-based filtering across different
continual learning settings. Each entry denotes the number of trajectories
satisfying the IDM reliability criterion over the total trajectories. Timeout rollouts are excluded before reliability scoring.
}
\label{tab:idm_reliable_ratio}
\end{table*}

%% file: tables/table10.tex
\begin{table*}[!ht]
\centering
\small

\RIFARTable
\begin{tabularx}{\textwidth}{cc|>{\centering\arraybackslash}X>{\centering\arraybackslash}X>{\centering\arraybackslash}X|>{\centering\arraybackslash}X>{\centering\arraybackslash}X>{\centering\arraybackslash}X}
\toprule
\multirow{2}{*}{\textbf{Stage}}
&
\multirow{2}{*}{\textbf{Model Pair}}
&
\multicolumn{3}{c|}{\cellcolor{FixedReplayFill}\textcolor{SuiteObjectInk}{\textbf{D$-$}}}
&
\multicolumn{3}{c}{\cellcolor{DriftFill}\textcolor{SuiteGoalInk}{\textbf{D$(+)$}}}
\\
\cmidrule(lr){3-5}
\cmidrule(lr){6-8}
&
&
\cellcolor{FixedReplayFill}\textbf{Mean}
&
\cellcolor{FixedReplayFill}\textbf{Max}
&
\cellcolor{FixedReplayFill}\textbf{Tail}
&
\cellcolor{DriftFill}\textbf{Mean}
&
\cellcolor{DriftFill}\textbf{Max}
&
\cellcolor{DriftFill}\textbf{Tail}
\\
\midrule
\multirow{4}{*}{Task0}
& M6 / Base & 0.4702 & 1.0278 & 0.9986 & 0.4622 & 1.2278 & 0.9851 \\
& M7 / Base & 0.8844 & 1.5086 & 1.3399 & 0.8732 & 1.5630 & 1.3642 \\
& M8 / Base & 1.0830 & 1.8404 & 1.6377 & 0.9980 & 1.7454 & 1.5116 \\
& M9 / Base & 1.2368 & 2.1180 & 1.9027 & 1.1925 & 2.1393 & 1.8564 \\
\midrule
\multirow{4}{*}{Task1}
& M6 / Base & 0.5246 & 1.3363 & 0.9653 & 0.4849 & 1.0559 & 0.8336 \\
& M7 / Base & 0.9532 & 2.3010 & 1.8192 & 0.9005 & 2.1672 & 1.7005 \\
& M8 / Base & 1.1439 & 2.2692 & 1.7995 & 1.0640 & 2.1126 & 1.6757 \\
& M9 / Base & 1.1319 & 1.9630 & 1.6588 & 1.0826 & 1.9638 & 1.5919 \\
\midrule
\multirow{4}{*}{Task2}
& M6 / Base & 0.7299 & 2.0123 & 1.5149 & 0.6481 & 1.4927 & 1.1938 \\
& M7 / Base & 1.3450 & 3.1758 & 2.3791 & 1.0817 & 2.0187 & 1.6307 \\
& M8 / Base & 1.4428 & 3.1799 & 2.5013 & 1.4083 & 3.0468 & 2.4145 \\
& M9 / Base & 1.4861 & 3.0998 & 2.5764 & 1.4444 & 2.9463 & 2.4679 \\
\midrule
\multirow{4}{*}{Task3}
& M6 / Base & 0.6780 & 1.8929 & 1.4150 & 0.6557 & 1.8331 & 1.3720 \\
& M7 / Base & 0.9599 & 2.5190 & 1.8091 & 0.8878 & 2.2653 & 1.6039 \\
& M8 / Base & 1.0848 & 2.7887 & 2.0037 & 1.0042 & 2.7010 & 1.8623 \\
& M9 / Base & 1.1851 & 3.0954 & 2.2025 & 1.1494 & 2.9663 & 2.0481 \\
\midrule
\multirow{4}{*}{Task4}
& M6 / Base & 0.8072 & 3.0531 & 1.9174 & 0.5707 & 1.5384 & 1.1316 \\
& M7 / Base & 0.8777 & 2.2010 & 1.5212 & 0.7995 & 1.1684 & 1.2225 \\
& M8 / Base & 1.2093 & 2.8327 & 2.1253 & 1.1053 & 2.3763 & 1.8339 \\
& M9 / Base & 1.2180 & 2.6270 & 2.0648 & 1.1831 & 2.5046 & 1.9428 \\
\midrule
\multirow{4}{*}{Task5}
& M6 / Base & 0.5987 & 1.7364 & 1.3820 & 0.5303 & 1.3084 & 1.0716 \\
& M7 / Base & 1.0681 & 2.7923 & 2.1518 & 0.9932 & 2.5452 & 1.9652 \\
& M8 / Base & 1.0112 & 2.4939 & 1.9570 & 0.9932 & 2.4285 & 1.8829 \\
& M9 / Base & 0.9194 & 2.1695 & 1.8226 & 0.8871 & 2.0902 & 1.7546 \\
\midrule
\multirow{3}{*}{Task6}
& M7 / M6 & 0.5947 & 1.0402 & 0.9046 & 0.5608 & 0.9951 & 0.8512 \\
& M8 / M6 & 0.8979 & 1.5276 & 1.3767 & 0.8511 & 1.6065 & 1.3636 \\
& M9 / M6 & 1.0569 & 1.7464 & 1.5407 & 1.0127 & 1.6696 & 1.5016 \\
\midrule
\multirow{2}{*}{Task7}
& M8 / M7 & 0.9610 & 2.1996 & 1.6654 & 0.8554 & 2.1114 & 1.6159 \\
& M9 / M7 & 1.2095 & 2.3265 & 2.0775 & 1.1226 & 2.0307 & 1.8673 \\
\midrule
Task8
& M9 / M8 & 0.7177 & 1.7683 & 1.3203 & 0.6572 & 1.3997 & 1.1490 \\
\bottomrule
\end{tabularx}
\caption{
Detailed action drift comparison across continual learning stages.
The model pair denotes the two models used to calculate behavioral drift.
D$-$ denotes replay without drift-aware selection, while D$(+)$ denotes replay
with drift-aware selection. Mean, Max, and Tail represent the average drift, maximum drift, and the average
drift of the top 20\% highest-drift samples, respectively.
}
\label{tab:detailed_drift}
\end{table*}

%% file: tables/table12.tex
\begin{table*}[t]
\centering

\RIFARTable
\renewcommand{\arraystretch}{1.1}
\setlength{\tabcolsep}{6pt}
\begin{tabular*}{\linewidth}{@{\extracolsep{\fill}}llccccccccc@{}}
\toprule
\rowcolor{TableHeader}
\textbf{Dataset}
&
\textbf{Setting}
&
\textbf{Task0}
&
\textbf{Task1}
&
\textbf{Task2}
&
\textbf{Task3}
&
\textbf{Task4}
&
\textbf{Task5}
&
\textbf{Task6}
&
\textbf{Task7}
&
\textbf{Task8}
\\
\midrule
\multirow{4}{*}{\textcolor{SuiteGoalInk}{Goal}}
& Model6
&4&1&3&2&3&1&--&--&--
\\
& Model7
&2&1&0&5&3&3&6&--&--
\\
& Model8
&6&1&5&8&2&10&2&2&--
\\
& Model9
&6&2&3&3&2&10&3&3&2
\\
\midrule
\multirow{4}{*}{\textcolor{SuiteObjectInk}{Object}}
& Model6
&3&3&1&2&4&2&--&--&--
\\
& Model7
&6&3&3&3&3&3&3&--&--
\\
& Model8
&5&3&5&3&4&10&8&5&--
\\
& Model9
&10&10&10&6&10&10&10&6&5
\\
\midrule
\multirow{4}{*}{\textcolor{SuiteSpatialInk}{Spatial}}
& Model6
&2&5&2&2&4&4&--&--&--
\\
& Model7
&4&3&4&3&3&5&4&--&--
\\
& Model8
&3&2&3&5&6&10&6&4&--
\\
& Model9
&2&5&5&3&7&10&5&2&2
\\
\bottomrule
\end{tabular*}
\caption{
Overlap of selected trajectories between replay1 and replay2 after the final
replay selection process. Each entry denotes the number of shared trajectories
between the two replay sets.
}
\label{tab:replay_overlap}
\end{table*}